ARTICLE

# Fabrication of Complex Soft Robotics: Challenges, Solutions, and Optimization

Mohammed Abboodi;

University of Ottawa; mabbo103@uottawa.ca ; https://orcid.org/0009-0002-4645-0026

**Abstract:** Manufacturing remains a major challenge in the development of complex soft pneumatic actuators, where geometric fidelity must be achieved without compromising compliance, structural integrity, or airtightness. This study introduces a systematic manufacturing framework in which a single complex actuator architecture is used as a common benchmark to evaluate heat-shrink forming, silicone casting, powder- and liquid-based additive manufacturing, and fused deposition modeling (FDM). The processes were assessed using the same criteria: geometric fidelity, flexibility, structural integrity, airtightness, manufacturing complexity, and post-processing requirements. A key contribution of the study is the use of observed failure mechanisms to distinguish inherent process limitations from defects that can be corrected through process development, providing a direct basis for manufacturing process selection. Heat-shrink forming, casting, powder-based processes, and the investigated digital light processing (DLP) route were limited primarily by geometric conformity, bonded interfaces, inaccessible residual material, and material and post-processing constraints, respectively. FDM was retained because its dominant defects could be progressively reduced without simplifying the actuator architecture. The results further showed that pneumatic sealing depends strongly on extrusion-path architecture rather than wall thickness alone, while support-free geometry is important for enclosed pneumatic structures. These findings provide practical manufacturing guidance for the design and fabrication of complex, flexible, and airtight soft pneumatic actuators.

## 1. Introduction

Soft pneumatic actuators have become an important actuation technology in soft robotics because they combine structural compliance, low weight, and large deformation with the ability to generate a wide range of motions [1], [2], [3]. Through appropriate design of pneumatic chambers, wall geometry, and structural constraints, these actuators can produce bending [4],[5], extension [6], contraction [7], twisting [8], [9], and combined motions [10], supporting applications in soft grippers [11], wearable robots[5], [12], rehabilitation devices [13], and continuum robotic systems. As actuator architectures become more complex, however, their successful implementation increasingly depends on the manufacturing process [9]. Thin deformable walls, enclosed pneumatic chambers, narrow internal passages, and intricate geometric features must be fabricated accurately while maintaining sufficient flexibility, mechanical integrity, and airtightness. Manufacturing is therefore not simply a final step in actuator development; it directly influences whether the intended pneumatic mechanism can operate reliably.

Heat-shrink forming has been reported as a rapid fabrication route for pneumatic actuators [14], [15]; however, rigid end closures and difficulties in achieving consistent sealing can reduce structural compliance and pneumatic reliability, particularly as geometric complexity increases [16]. Silicone molding and casting remain among the most widely used methods for fabricating soft pneumatic actuators because they can produce highly compliant structures using relatively simple equipment [1], [9], [17]. These methods are effective for many conventional actuator geometries[4], [18], but their limitations become more pronounced as geometric complexity increases [9]. Enclosed structures may require multi-part molds, while narrow cavities and thin walls make material filling and demolding increasingly difficult [9], [19]. Complex actuators may also need to be fabricated as several individual components that are subsequently bonded to form a closed pneumatic structure. These interfaces introduce additional manufacturing steps and can become weak regions where leakage or mechanical failure occurs. Such limitations become particularly important when the actuator must undergo large and repeated deformations while maintaining a sealed pneumatic chamber.

Additive manufacturing offers an alternative route by producing components directly from three-dimensional models and reducing the dependence on molds and secondary assembly [20], [21]. Internal channels, pneumatic chambers, and structural features can potentially be integrated into a single component, greatly expanding the geometric freedom available to soft robotic designers [9], [10]. Several additive-manufacturing approaches have therefore been explored for soft pneumatic structures, including powder-based processes [22], vat photopolymerization [23], [24], direct ink writing[25], and material extrusion[26]. These techniques have demonstrated considerable potential for producing increasingly complex actuator architectures.

However, the ability to reproduce a complex geometry does not alone define a successful manufacturing process for a soft pneumatic actuator. The fabricated structure must also remain compliant and maintain pneumatic integrity during large deformation. Each manufacturing route introduces different challenges. Powder-based methods require residual material to be removed from enclosed cavities and narrow passages[10], [16]. Resin-based processes can provide high geometric resolution but depend strongly on the mechanical properties of the cured photopolymer and generally require draining, washing, and post-curing. Material-extrusion processes provide direct fabrication using flexible thermoplastics, but their performance is highly sensitive to material condition, extrusion stability, interlayer bonding, deposition accuracy, and void formation. Even small discontinuities that have little effect on the external appearance of a printed component may create continuous leakage paths through a pneumatic wall.

Despite substantial progress in soft actuator fabrication, manufacturing is often treated as a supporting step in the development of a particular actuator or robotic system. A manufacturing technique is typically selected, adjusted until a functional prototype can be produced, and then followed by mechanical characterization, modeling, or control. However, systematic manufacturing-focused studies that address process selection, failure mechanisms, and process optimization for complex soft pneumatic actuators remain scarce. Consequently, there remains a need for a more systematic approach that begins with the manufacturing requirements of a complex pneumatic structure, identifies why fundamentally different fabrication routes succeed or fail, and distinguishes limitations inherent to a manufacturing process from defects that can be corrected through process optimization. This distinction is particularly important for actuators that simultaneously require thin flexible walls, enclosed chambers, narrow internal passages, complex surface features, and reliable airtightness.

The present study addresses this need by treating manufacturing as the primary research problem rather than as a supporting fabrication step. A previously developed complex soft pneumatic actuator architecture [16] was used as a representative manufacturing challenge. Heat-shrink forming, silicone casting, powder- and liquid-based additive manufacturing, and material extrusion were evaluated according to their ability to reproduce the required geometry while maintaining flexibility, structural integrity, and airtightness. The study investigates the following research question: Which manufacturing technique, and under what processing conditions, can reliably produce geometrically complex, flexible, and airtight soft pneumatic actuators? Manufacturing routes constrained by fundamental process limitations were identified and excluded, whereas correctable defects were analyzed and progressively addressed through process development and optimization.

## 2. Method

A structured experimental methodology was used to systematically evaluate and develop manufacturing processes for geometrically complex soft pneumatic actuators. A representative actuator architecture reported previously [16] was selected because it combines thin flexible walls, enclosed pneumatic chambers, narrow internal passages, and folded features that present significant manufacturing challenges. The investigated processes were evaluated using a common set of criteria: geometric fidelity, flexibility, structural integrity, airtightness, manufacturing complexity, and post-processing requirements.

The study followed four main stages: process evaluation, baseline fabrication, failure analysis, and process improvement and validation. In the first stage, heat-shrink forming, silicone casting, and additive-manufacturing processes were examined against the geometric and functional requirements of the actuator. This provided a consistent basis for comparing the capabilities, limitations, and manufacturing requirements of each process.

Baseline actuators were then fabricated where applicable, and the principal manufacturing defects were identified. These included geometric distortion, incomplete wall formation, voids, weak interfaces, trapped material, and pneumatic leakage. Each defect was analyzed to determine whether it resulted from an inherent limitation of the manufacturing process or from an adjustable process condition.

Process improvement focused on the manufacturing variables associated with the observed defects. For FDM, development followed a hierarchical sequence involving material conditioning, extrusion stability, support-free design, build-plate adhesion, deposition parameters, wall architecture, and cooling conditions. The developed process was subsequently evaluated using actuator models with different geometries and assessed in terms of geometric integrity, flexibility, and pneumatic performance.

# 3. Manufacturing

## 3.1 Heat-Shrink-Based Manufacturing

Heat-shrink-based manufacturing was investigated as a simple and rapid method for fabricating geometrically complex soft pneumatic actuators. The method uses a thermoplastic tube that contracts when heated and conforms to the surface of an internal mold. In principle, this approach can reduce the number of fabrication and assembly steps while producing a thin and continuous actuator wall. The process was evaluated to determine whether heat-shrink tubing could reproduce deep geometric features while maintaining flexibility, structural integrity, and airtightness. The manufacturing procedure consisted of mold fabrication, thermal forming, mold removal, and end-cap integration.

### 3.1.1 Manufacturing Process

The mold was designed to reproduce the corrugated surface of the actuator (Figure 1a and 1b). A relaxed 3M FP-301 heat-shrink tube was placed around the mold and then heated until it contracted toward the mold surface. Because the final actuator geometry depended directly on the stability of the mold, selecting a suitable mold material was essential.

The first molds were fabricated from acrylonitrile butadiene styrene (ABS). However, the temperature reached during forming exceeded the glass-transition temperature of ABS, approximately 105 °C, causing the mold to deform and reducing the accuracy of the formed structure. Polycarbonate was therefore selected as an alternative because of its higher glass-transition temperature of approximately 150 °C. This material maintained its shape more effectively during heating and resisted the forces produced by the contracting tube.

Three heating methods were evaluated: a heat gun, an oven, and boiling water. Uniform heating was necessary to avoid uneven contraction and distortion of the actuator wall. The heat gun produced localized heating, which made the forming process difficult to control. Boiling water provided the most uniform and predictable heat distribution because the entire mold–tube assembly was exposed to nearly the same temperature. It was therefore selected as the preferred heating method. The assembly was immersed in boiling water until the tube had contracted around the mold and was then cooled before demolding.

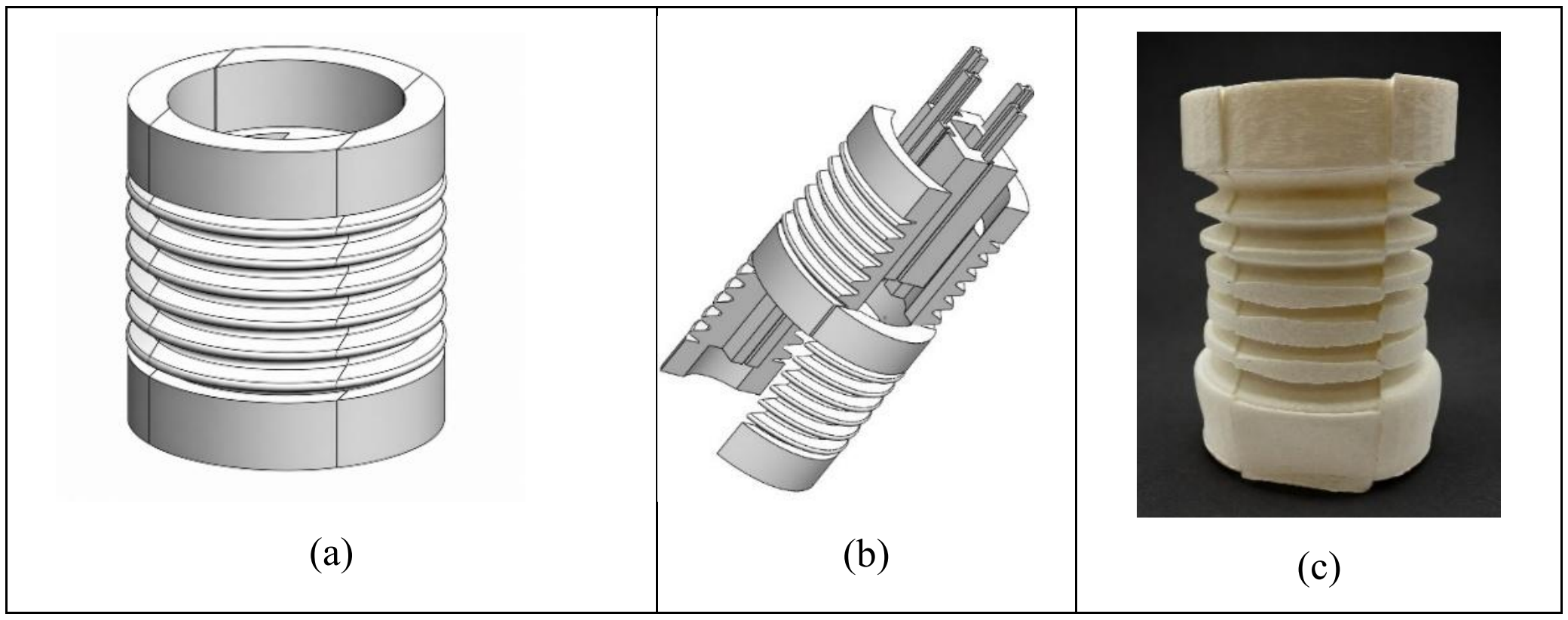

**Figure1.** Soft actuator heat-shrink mold: (a) complete mold assembly, (b) disassembled mold parts (c) mold after heat exposure

Removing the mold was challenging because the enclosed corrugated structure prevented the use of a conventional one-piece mold. Pulling a solid mold from the formed tube could stretch, tear, or permanently deform the thin wall. To overcome this problem, a modular mold was developed from several interlocking sections, as shown in Figure 1b and Figure 1c. The central section was removed first, creating enough internal space for the surrounding sections to move inward and be removed individually. I-shaped keys were used to align and secure the mold sections during heating while still allowing the mold to be disassembled after forming.

Upper and lower end caps were then added to close the structure and create a pneumatic chamber. The ABS caps were positioned at both ends of the formed tube, and only the regions surrounding the caps were reheated. This localized heating caused the tube to contract tightly around the caps while limiting further thermal exposure to the actuator body. The assembled mold, removable mold sections, and thermally formed structure are shown in Figure 1c.

### 3.1.2 Limitations and Manufacturing Outcome

The main limitation of the process was its inability to reproduce the deep valleys of the corrugated actuator geometry. The selected FP-301 tubing had a maximum shrink ratio of 4:1, which was not sufficient to conform fully to the mold surface. During heating, the tube first contacted the outer peaks of the mold. These contact points then acted as anchors, causing the remaining material to contract across the gaps between adjacent peaks rather than move into the recessed valleys. As a result, the tube bridged over the deeper features and did not reproduce the intended geometry accurately.

Applying additional heat did not improve conformity. Once the tubing reached its maximum contraction, further heating damaged the material and eventually caused tearing. The loss of geometric accuracy was therefore not caused only by uneven heating. It resulted mainly from the limited shrink ratio of the tubing relative to the depth and spacing of the mold features.

The rigid ABS end caps introduced another limitation. Although they provided a practical means of closing the actuator, their stiffness reduced the overall flexibility of the structure and created rigid regions near the actuator ends. This is undesirable in soft pneumatic actuators, where the structure should deform freely without significant resistance from rigid components.

Complete airtightness was also not achieved consistently. The experiments confirmed the presence of air leakage. Leakage may have originated from damage to the tubing during heating or mold removal, or the interfaces between the tubing and the end caps. Because these possibilities were not tested separately, no single leakage mechanism can be confirmed from the available results.

Overall, heat-shrink forming offered a relatively simple and rapid method for producing tubular structures. The use of boiling water improved heating uniformity, while the modular mold reduced the risk of damage during demolding. However, the method could not provide the geometric accuracy, flexibility, and airtightness required for highly complex soft pneumatic actuators. It may be suitable for actuators with smooth surfaces or shallow geometric features, but the selected tubing and forming process were not suitable for designs containing deep corrugations and narrow recessed regions. The results show that the available shrink ratio was the main factor limiting the use of this manufacturing method for complex pneumatic structures.

## 3.2 Casting-Based Manufacturing

Casting was investigated as a potential method for producing complex soft pneumatic actuators because it can form highly flexible silicone components using relatively simple equipment. However, the actuator geometry examined in this study included enclosed chambers, narrow internal spaces, thin walls, and multiple interconnected features. The method was therefore evaluated based on its ability to reproduce this geometry while maintaining flexibility, mechanical integrity, and airtightness.

### 3.2.1 Manufacturing Process

A multi-part mold was designed to form the internal wall, external wall, and end caps of the actuator. The mold components were fabricated from polylactic acid using material-extrusion 3D printing, as shown in Figure 2a. Dividing the mold into separate parts made it possible to form the enclosed geometry and simplified mold removal after curing.

Dragon Skin silicone rubber (Smooth-On, Easton, PA, USA) was selected because of its elasticity and durability. The two-part silicone was mixed and placed in a vacuum chamber for 30 min to remove air bubbles introduced during preparation. This step was intended to reduce internal defects and improve the consistency of the cured material.

The internal and external walls were cast first by pouring the silicone into their corresponding mold cavities. The upper and lower caps were then formed and integrated using custom-designed molds. Silicone sealant was applied at the interfaces between the mold components to prevent uncured silicone from escaping during casting.

The assembly was allowed to cure at room temperature for 16 h. After curing, the mold was separated into four parts and removed gradually to reduce the risk of damaging the silicone walls and internal features. Figure 2b shows the fabricated prototype and the disassembled molds used during the casting process.

### 3.2.2 Limitations and Manufacturing Outcome

Although casting produced a complete silicone prototype, several problems limited its suitability for highly complex soft pneumatic actuators. The main challenges were weak bonding between separately cast components, the formation of internal defects, and inconsistent airtightness.

The actuator had to be fabricated from several components and subsequently assembled to create a closed pneumatic structure. The bonded interfaces formed mechanically weak regions and were prone to failure at internal pressures above approximately 200 kPa. This showed that the mechanical integrity of the actuator depended not only on the properties of the silicone but also on the quality and consistency of the bonded interfaces. As the number of separately cast components increased, these interfaces became potential failure locations.

Reliable airtightness was also difficult to achieve. Voids, small holes, and discontinuities were observed within the cast walls, allowing air to escape during pressurization. These defects were associated with gas trapped in the silicone and within the narrow mold cavities. Although vacuum degassing reduced some of the entrapped air, it did not completely prevent voids and discontinuities from forming.

Wall thickness strongly affected the outcome of the casting process. In this study, a wall thickness of approximately 5 mm was required to produce complete actuator walls with fewer defects. However, this thickness made the actuator bulky and reduced its flexibility. The process therefore introduced a clear trade-off. Increasing the wall thickness improved the likelihood of successful fabrication but also increased stiffness and resistance to deformation. Thinner walls provided greater flexibility but were more difficult to cast without incomplete filling, internal voids, or air leakage.

The enclosed geometry further complicated the process because the silicone had to flow through narrow spaces while trapped air escaped from the same regions. This made it difficult to produce thin, continuous, and defect-free walls throughout the actuator. Vacuum degassing alone could not resolve this problem because defect formation was also influenced by the accessibility and complexity of the mold cavities.

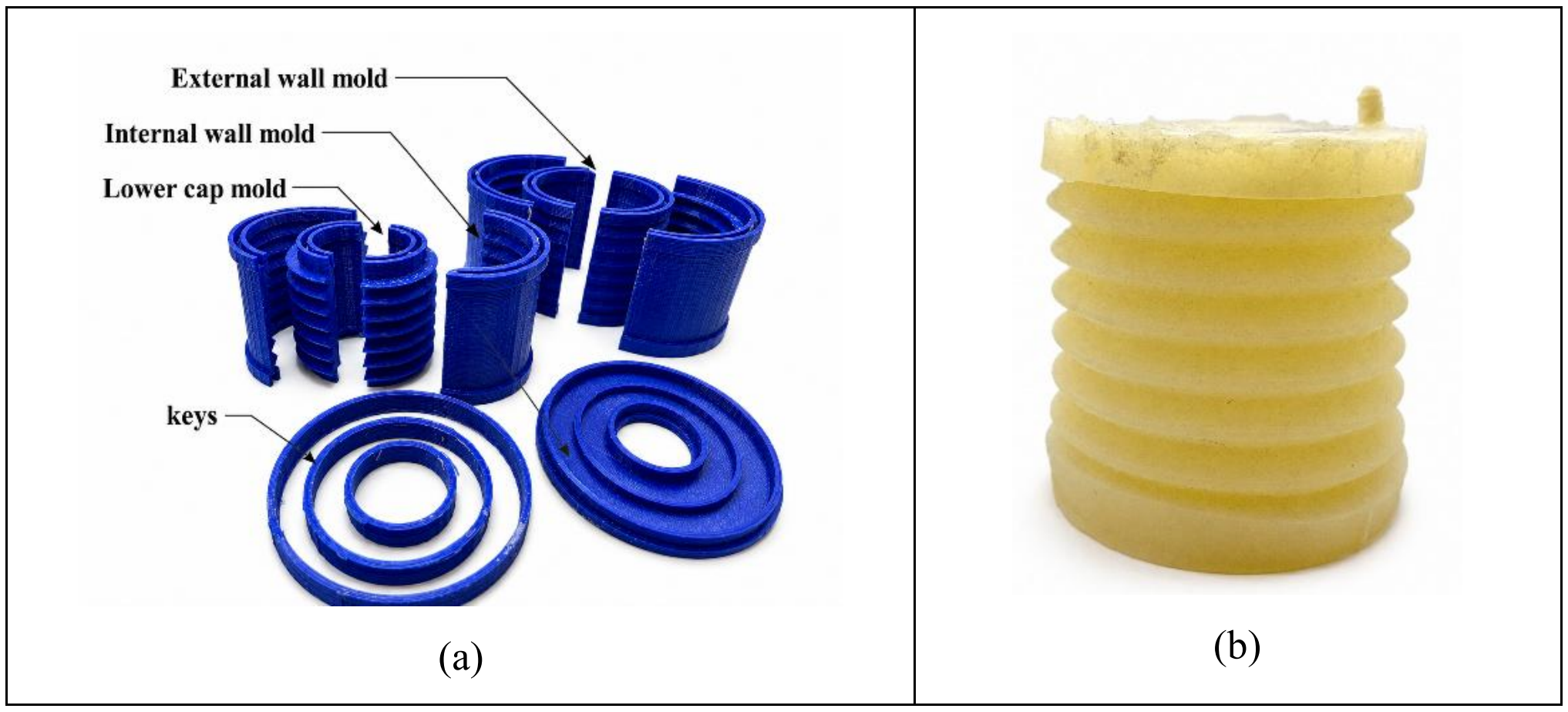

**Figure 2.** Fabrication by casting: (a) disassembled mold components used during the casting process, (b) cast prototype.

Overall, casting remains a practical method for soft pneumatic actuators with relatively simple geometries, limited assembly requirements, and sufficiently thick walls. However, it did not provide the combination of thin-wall fabrication, reliable bonding, flexibility, and airtightness required for the complex pneumatic structures examined in this study. The need for walls of approximately 5 mm also resulted in actuators that were too rigid and bulky. Casting was therefore not retained as the final manufacturing method, indicating the need for a process capable of producing the complete actuator geometry with fewer bonded interfaces and better control over thin-wall formation.

## 3.3 Additive Manufacturing

Additive manufacturing was investigated as an alternative to heat-shrink forming and casting because it can produce complex parts directly from three-dimensional models. Unlike conventional methods, which often require several molds, assembly steps, and bonded interfaces, additive manufacturing builds the component layer by layer. This makes it attractive for soft pneumatic actuators with enclosed chambers, thin walls, narrow air passages, and complex internal features.

However, printing these structures with flexible materials remains challenging. The selected process must reproduce the required geometry without blocking internal passages or damaging thin walls during post-processing. The material must also retain enough flexibility for pneumatic actuation. The additive-manufacturing methods considered in this study were therefore grouped according to their feedstock: powder-based, liquid-based, and filament-based processes. Each group was assessed based on its material compatibility and its ability to produce complex enclosed geometries.

### 3.3.1 Powder-Feedstock Additive Manufacturing

Powder-based additive manufacturing includes several processes with different operating principles and material capabilities. Among these methods, selective laser sintering (SLS) and Multi Jet Fusion (MJF) can process flexible thermoplastic polymers. Other powder-based techniques, including direct metal laser sintering, selective laser melting, and electron beam melting, are mainly used for metallic materials and were therefore not considered suitable for flexible pneumatic actuators.

In SLS, a carbon dioxide laser selectively fuses thermoplastic powder according to the cross-section of the digital model. After each layer is formed, the build platform is lowered and covered with a new layer of powder. The surrounding unfused powder supports the part during printing, allowing complex geometries to be produced without separate support structures.

MJF also uses a powder bed, but the material is fused by applying functional agents to selected regions and exposing the layer to infrared energy. This process can produce complex components and offers some control over the properties of the printed material.

Both SLS and MJF can form complex external geometries without conventional supports. However, the unfused powder must be removed after printing. This becomes difficult when the structure contains enclosed chambers, narrow passages, or internal cavities as small as approximately 1 mm. Powder can remain trapped inside these regions and interfere with the internal pneumatic network.

The main limitation of powder-based manufacturing was therefore not its ability to reproduce the external shape, but the difficulty of removing residual powder from enclosed internal features. Although SLS and MJF may be suitable for flexible components with open or accessible cavities, they were not selected for the highly complex pneumatic structures examined in this study. Their use would require internal passages large enough to allow complete powder removal after printing.

### 3.3.2 Liquid-Feedstock Additive Manufacturing

Liquid-based additive manufacturing produces three-dimensional components by curing liquid photopolymers or depositing printable inks layer by layer. Four processes were considered: direct ink writing (DIW), PolyJet printing (PolyJet), stereolithography (SLA), and digital light processing (DLP). These methods were assessed based on their ability to process flexible materials and produce thin walls, enclosed chambers, and narrow internal passages.

DIW builds structures by extruding shear-thinning inks through a nozzle. Although it can process soft materials, the deposited layers may deform before solidification. Nozzle clogging, slow printing speeds, and demanding post-processing also limit its use for highly complex pneumatic structures.

PolyJet printing deposits small droplets of liquid photopolymer and cures them using ultraviolet light. The process can produce detailed geometries, combine multiple materials, and use removable supports to form complex features. However, the high cost of the equipment and proprietary materials made it impractical for this study.

SLA and DLP both use ultraviolet light to cure photosensitive resin. SLA uses a scanning laser to cure the resin point by point, whereas DLP projects an image and cures an entire layer at once. Both methods allow uncured resin to be drained and washed from enclosed cavities, making them more suitable for complex internal geometries than powder-based processes. DLP was therefore selected for experimental evaluation.

*3.3.2.1 DLP Manufacturing Process*

Preliminary manufacturing trials were conducted using an Elegoo Mars 2 printer and two flexible photopolymer resins: Liqcreate Premium Flex, with a Shore hardness of 63A, and Liqcreate Flexible-X, with a Shore hardness of 55A. Before printing, the resin was shaken to ensure a uniform mixture and then placed in a vacuum chamber for 15 min to remove air bubbles that could affect print quality. The main printing parameters, including layer thickness, exposure time, base-layer settings, lift height, lift and retract speeds, and light-off delay, were adjusted experimentally. The selected settings for the two resins are summarized in Table 1.

Printing hollow structures created suction forces as each cured layer separated from the resin vat. These forces could deform the thin walls or detach the component from the build platform. To reduce this effect, the models were printed at an orientation of 45°. Vent holes were also added to allow air and uncured resin to move through the internal cavities during printing. After printing, the holes were sealed using a custom mold and additional ultraviolet curing.

The printed components were washed in 95% ethanol to remove uncured resin from the external surfaces and internal passages. They were then exposed to ultraviolet light for 45 min to complete the curing process. This procedure successfully produced a complete flexible pneumatic actuator.

**Table 1:** Optimal printing parameters for DLP printer

| **Resin** | **Layer thickness** | **Exposure time (s)** | **Base layer count** | **Base layer exposure** | **Lift Height(mm)** | **Lift / retract speed (mm/min)** | **Light off-delay / bottom (s)** |
|---|---|---|---|---|---|---|---|
| **Premium Flex** | 50 µm | 5.5 | 5 | 60 | 10 | 60 / 90 | 2/ 3 |
| **Flexible-X** | 50 µm | 17 | 2 | 60 | 10 | 90 / 120 | 1 / 3 |

*3.3.2.2 Limitations and Manufacturing Outcome*

Although DLP successfully reproduced the complex actuator geometry, several limitations prevented its selection as the final manufacturing method. Flexible photopolymer resins were considerably more expensive than the other materials examined in this study. This cost becomes important when producing large components, multiple prototypes, or several design variations.

The process also required careful material handling, suitable ventilation, and several post-processing steps. These included draining and washing the internal cavities, sealing the vent holes, and completing the ultraviolet curing. Each additional step increased the manufacturing time and introduced another opportunity for defects, particularly in narrow or difficult-to-access internal passages.

The main limitation was the mechanical behavior of the cured resins. Their elongation was below 150%, which restricted the large and repeated deformations required for highly flexible pneumatic actuation. Although the process could reproduce the intended geometry, geometric accuracy alone was not sufficient. The printed material also needed to withstand repeated deformation without cracking, tearing, or losing its mechanical properties.

Overall, DLP was more effective than casting and powder-based printing in producing detailed and enclosed geometries. The use of a 45° printing orientation, vent holes, optimized exposure settings, and controlled post-curing enabled the successful fabrication of a complete prototype. However, the high material cost, extensive post-processing, ventilation requirements, and limited resin elongation reduced its suitability for highly deformable soft pneumatic actuators. DLP was therefore not retained as the final manufacturing route.

### 3.3.3 Fused Deposition Modeling

Fused deposition modeling (FDM) was investigated as the principal manufacturing route because it can produce complex pneumatic structures directly from three-dimensional models. Unlike casting, FDM can integrate thin walls, enclosed chambers, internal air channels, and structural constraints into a single component. This reduces the need for molds, assembly steps, and bonded interfaces. Thermoplastic polyurethane (TPU) also provides the flexibility and large deformation required for soft pneumatic actuation.

Producing airtight structures by FDM, however, is more difficult than printing conventional flexible components. Small gaps between adjacent extrusion lines, weak interlayer bonding, moisture-related porosity, or inconsistent material flow can create leakage paths. At the same time, increasing material deposition to improve airtightness can make the actuator too stiff and restrict its deformation. Therefore, the manufacturing problem is not simply to print a complete geometry, but to establish a process window that balances airtightness, flexibility, dimensional accuracy, and manufacturing reliability.

*3.3.3.1 Selection of the Extrusion System*

FDM printers generally use either a direct-drive or Bowden extrusion system. In a direct-drive system, the extruder motor is mounted on the print head, creating a short filament path between the drive mechanism and the nozzle. This arrangement is commonly preferred for flexible materials because it reduces filament compression and buckling before extrusion.

Initial manufacturing trials were conducted using a Prusa i3 MK3S direct-drive printer. However, long and geometrically complex prints were affected by vibration, layer misalignment, filament entanglement, and repeated nozzle clogging. The mass of the moving print head also reduced positional accuracy during rapid changes in direction. These problems lowered the reliability of long printing operations and prevented the consistent production of complete pneumatic structures. The tested direct-drive system was therefore not retained for further development.

In a Bowden system, the extruder motor is mounted on the printer frame, and the filament is guided to the print head through a flexible tube. The lower print-head mass improves motion stability and positional accuracy, which is useful when printing repeated folds, narrow channels, and thin walls. However, the longer filament path increases friction and makes flexible TPU more likely to compress, buckle, or kink before reaching the nozzle.

Despite these challenges, the Bowden configuration provided better motion stability for the complex geometries. A controlled manufacturing framework was therefore developed using the Ultimaker 3 Extended and Ultimaker S3 printers. The resulting controlled-environment printing setup is shown in Figure 3.

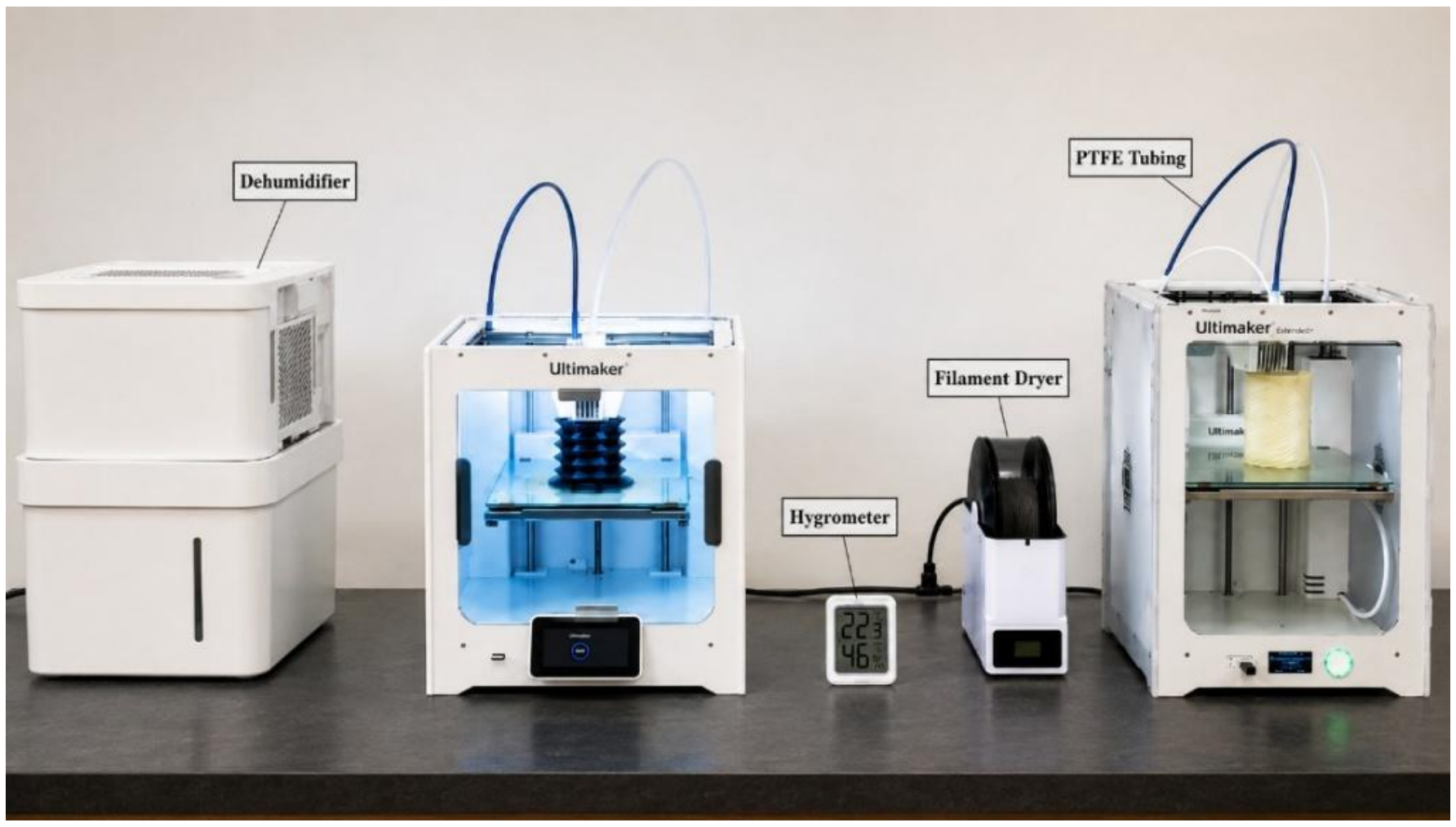

**Figure 3.** Additive manufacturing and environmental control setup used for soft pneumatic actuator fabrication.

*3.3.3.2 Moisture Control and Filament Conditioning*

TPU moisture had a noticeable effect on extrusion stability and the quality of the printed pneumatic walls. During preliminary printing trials, moist filament was associated with irregular material flow, visible bubbles and pores, and less consistent bonding between deposited lines. These defects are consistent with the hygroscopic behavior of TPU. When absorbed moisture reaches the heated nozzle, it vaporizes within the polymer melt and disturbs the continuity of extrusion. The resulting fluctuations in material flow can produce local under-extrusion and voids, while reducing the contact and fusion between adjacent extrusion paths and successive layers. Similar moisture-related effects on TPU extrusion and bonding have been reported in previous studies [10], [27].

A two-stage conditioning procedure was therefore adopted. The relative humidity around the printer was maintained below 20% to limit moisture uptake during long printing operations, and the TPU filament was dried for 5 h at 50 °C before printing. Silica gel was also used during conditioning and storage to maintain a dry environment. After conditioning, extrusion became more stable and the occurrence of visible pores and discontinuities in the printed walls was reduced. The controlled printing environment is shown in Figure 3.

Moisture conditioning was established before optimization of the main printing parameters because changes in temperature, speed, or flow rate could not reliably compensate for unstable material delivery caused by moisture. The observations therefore link filament condition directly to wall porosity, extrusion stability, and interlayer bonding, all of which influence the continuity and airtightness of thin pneumatic walls. Moisture control should consequently be treated as part of the FDM manufacturing process rather than solely as a filament-storage precaution.

*3.3.3.3 Stabilization of TPU Extrusion in a Bowden System*

After controlling moisture, the next challenge was to maintain a stable flow of TPU through the Bowden extrusion system. In this configuration, the filament travels through a relatively long tube before reaching the nozzle. Because TPU is soft and flexible, it can compress, buckle, kink, or coil along this path. Friction inside the tube also resists filament movement and delays the response to extruder commands. These effects can cause irregular material flow, local under-extrusion, or complete interruption of the printing process.

A four-step procedure was developed to improve extrusion stability. First, the extruder tension was adjusted gradually. Excessive tension compressed and deformed the filament, increasing the risk of blockage. Insufficient tension allowed the drive mechanism to slip, resulting in inconsistent feeding and under-extrusion. The tension was therefore increased in small steps until the filament moved continuously without visible deformation or slipping.

Second, resistance along the filament path was reduced. The standard Bowden tube was replaced with a shorter, low-friction polytetrafluoroethylene (PTFE) tube. Reducing the length and friction of the feed path limited filament compression and lowered the risk of buckling, kinking, and coiling before the material reached the nozzle.

Third, nozzle cleaning was included as a regular part of the manufacturing procedure. Flexible filaments were particularly sensitive to partial blockage during long printing operations. A cold-pull method was used to remove loose residue, while hot cleaning was applied to clear more persistent deposits. Keeping the nozzle clean helped prevent gradual reductions in material flow and local defects in the printed walls.

Finally, retraction was disabled. Repeated retraction stretches and compresses the viscoelastic TPU inside the Bowden tube, leading to unstable pressure at the nozzle and delayed extrusion after travel movements. Disabling retraction improved the continuity of material flow. The resulting stringing and oozing were controlled through the printing temperature, speed, and travel settings discussed in the following sections.

*3.3.3.4 Support-Free Manufacturing*

Support materials were initially evaluated for overhanging features and enclosed internal regions. Polyvinyl alcohol (PVA) was tested as a soluble support, but its use with TPU increased stringing and disrupted wall continuity. After dissolution, gaps and surface defects were also observed, which compromised pneumatic sealing. TPU-based supports showed better material compatibility but were difficult to remove from narrow internal spaces. Mechanical removal often damaged the thin walls, resulting in small holes that could not be repaired reliably. These observations showed that, for enclosed pneumatic structures, support removal can itself become a source of wall damage and leakage.

A support-free design was therefore investigated. Printing trials showed that overhangs below approximately 30° from the build plane were more susceptible to sagging and local wall weakening. In contrast, overhangs of approximately 30° or greater provided sufficient support from the preceding layers to maintain wall continuity and complete the enclosed geometry without internal supports. The 30° criterion was therefore derived from the observed printing behavior of the investigated actuator rather than adopted as a general FDM design rule.

The identified value is specific to the TPU material, printer configuration, actuator geometry, and processing conditions used in this study and should not be interpreted as a universal overhang limit. The broader implication is that internal accessibility and support requirements should be considered during the design of enclosed pneumatic structures. When internal supports cannot be removed without risking damage to the pressure boundary, self-supporting geometry provides a more reliable manufacturing strategy and reduces dependence on post-processing. In this context, support-free design becomes part of the actuator manufacturing strategy rather than a secondary consideration after printing.

*3.3.3.5 Build-Plate Adhesion*

Reliable first-layer adhesion was necessary because even minor detachment or warping could cause the entire print to fail. This was particularly important for flexible TPU structures, which can deform in response to nozzle movement and temperature changes.

A washable glue stick was applied to the build surface to improve initial adhesion. A brim was also added around the component to increase the contact area and reduce the risk of warping or detachment. The initial printing speed, first-layer temperature, and build-plate temperature were adjusted together to provide a stable foundation without excessive spreading or deformation. The final adhesion settings are reported in Table 2.

*3.3.3.6 Selection and Optimization of Printing Process Parameters*

The main printing parameters were adjusted iteratively by examining material flow, wall continuity, geometric accuracy, surface quality, and airtightness. Rather than treating each parameter independently, the optimization considered their combined influence on the formation of continuous and flexible walls.

*Extrusion temperature:* The temperature study began at 260 °C, after which the temperature was adjusted in increments of 5 °C. A printing temperature of 235 °C provided the best balance between material flow and geometric stability for the tested TPU materials and printers.

Below approximately 225 °C, the material became too viscous to flow consistently through the nozzle. This increased the risk of clogging, under-extrusion, and weak bonding between successive layers. The resulting gaps reduced wall strength and created paths for pneumatic leakage.

Above approximately 240 °C, the filament became excessively fluid. Oozing and stringing reduced surface quality and dimensional accuracy, while slower solidification caused deformation in unsupported or detailed regions. The suitable temperature range was therefore limited by the need to achieve strong layer bonding without losing control of the printed geometry.

*Printing speed:* The printing speed was initially set to 40 mm/s and reduced in increments of 5 mm/s. A speed of 15 mm/s provided the most consistent balance between layer placement, bonding, and manufacturing time.

At higher speeds, the interaction time between adjacent extrusion lines was reduced, limiting thermal bonding and increasing the likelihood of gaps or delamination. Rapid nozzle movement also reduced positional accuracy around narrow folds and curved features.

Speeds below approximately 10 mm/s did not provide a clear improvement. The longer residence time of the filament inside the heated nozzle increased material accumulation and produced surface irregularities. A speed of 15 mm/s was therefore selected as the most suitable value.

*Material flow rate:* The material flow rate was adjusted from an initial value of 100%. A flow rate of 110% provided sufficient material to close the interfaces between adjacent extrusion lines without causing excessive accumulation.

At flow rates of approximately 105% or lower, under-extrusion produced thin regions and discontinuities in the walls. At values above approximately 125%, excess material caused blobs, oozing, and dimensional errors. These defects reduced the accuracy of the internal passages and could interfere with the movement of nearby structural features.

Layer height and overlap: A layer height of 0.1 mm was selected to balance geometric resolution, interlayer bonding, and manufacturing time. Smaller layer heights greatly increased the printing time and the risk of nozzle blockage without providing a clear improvement in actuator performance. Larger layer heights reduced geometric accuracy and weakened the interfaces between successive layers.

An overlap of 20–25% provided sufficient bonding between adjacent printed regions. Values below approximately 15% increased the formation of gaps and weak interfaces, whereas values above approximately 30% caused material accumulation and reduced dimensional accuracy.

*Nozzle diameter and line width:* A nozzle diameter of 0.4 mm was selected for the final process. The smaller 0.25-mm nozzle clogged frequently when printing flexible TPU, whereas the 0.8-mm nozzle provided more stable material flow but could not reproduce the thin walls and detailed internal features required by the actuator geometry. A line width of 0.32 mm was therefore used with the 0.4-mm nozzle to maintain geometric accuracy while producing thin and flexible walls.

*Line width and wall-line count:* Line width had a pronounced effect on pneumatic sealing when considered together with the number of wall lines. Experimentally, a 1.6-mm wall produced using two 0.8-mm extrusion lines showed poorer airtightness than a thinner 0.96-mm wall produced using three 0.32-mm lines. Thus, increasing the nominal wall thickness by using wider extrusion lines did not improve sealing under the investigated conditions. One- and two-line walls were more prone to leakage, whereas increasing the number of wall lines improved wall continuity. However, four or five lines also increased stiffness, material consumption, and printing time. Three 0.32-mm lines therefore provided the most suitable balance between airtightness, flexibility, and manufacturing efficiency.

The experiments established the difference in sealing performance, but the internal leakage paths were not directly identified. One possible explanation is that a defect or incomplete bond associated with a wider extrusion line occupies a larger portion of the pressure boundary. With three narrower lines, a local defect in one deposited path does not necessarily extend through the full wall thickness because additional material paths remain between the internal chamber and the external surface. This may reduce the likelihood of forming a continuous leakage path. Because the internal defects were not directly visualized, this explanation is considered a proposed mechanism rather than an experimentally confirmed failure mode.

The experimental finding therefore shows that nominal wall thickness alone is not sufficient to define the airtightness of an FDM-manufactured pneumatic wall. Under the investigated conditions, the combination of narrower extrusion lines and an appropriate wall-line count provided better sealing than a thicker wall formed from fewer, wider lines. Line width and wall-line count should therefore be considered together when designing thin, flexible, and airtight pneumatic structures.

*Cooling conditions:* Cooling affected both geometric stability and interlayer bonding. A fan-speed range of 30–40% provided the most suitable balance for the tested structures.

At fan speeds below 30%, the deposited TPU remained soft for too long, increasing sagging and deformation around overhangs and narrow features. At speeds above 40%, rapid cooling reduced bonding between successive layers and increased the risk of delamination and air leakage. The final cooling settings are reported in Table 2.

The parameter study showed that airtight FDM fabrication is governed by the combined effects of material delivery, deposition conditions, wall geometry, and cooling rather than by any single printing parameter. Therefore, the optimized settings define an experimentally established process window for the investigated TPU materials, Bowden-based Ultimaker platforms, actuator geometry, and controlled printing environment. More importantly, FDM was retained because its dominant manufacturing limitations could be addressed through process optimization while preserving the integrated actuator architecture. This enabled geometric complexity, flexibility, airtightness, internal accessibility, and limited post-processing to be achieved within a single manufacturing route, providing the best overall match to the functional requirements of the actuator.

**Table 2:** Optimized printing settings

| Parameter | Value | Unit | Parameter | Value | Unit |
|---|---|---|---|---|---|
| **Quality** | | | | | |
| Nozzle diameter | 0.4 | mm | Build plate adhesion type | Brim | - |
| Layer height | 0.1 | mm | Brim width | 10 | mm |
| Initial layer height | 0.27 | mm | Brim only outside | Enable | - |
| Line width | 0.32 | mm | Brim speed | 20 | mm/s |
| Wall line count | 3 | - | Z hope speed | 20 | mm/s |
| **Infill** | | - | Optimize printing order | Disable | - |
| Infill density | 100% | - | Alternate extra wall | Disable | - |
| Infill Pattern | Zigzag | - | Infill pattern | line | - |
| **Material** | | | Acceleration control | Enable | - |
| Printing Temperature | 235 | °C | Infill Acceleration | 1500 | mm/s$^2$ |
| Initial layer printing temperature | 257 | °C | Wall ordering | Inside to outside | - |
| Build-plate temperature | 40 | °C | Wall Acceleration | 750 | mm/s$^2$ |
| Flow | 110 % | - | Jerk control | Enable | - |
| Infill Flow | 125% | - | Retraction | Disable | - |
| **Speed** | | | Support material | Disable | - |
| Print speed | 15 | mm/s | Avoid printed part when travel | Enable | - |
| Initial layer print speed | 11 | mm/s | Jerk control | Enable | - |
| Top/ Bottom speed | 12 | mm/s | Retraction | Disable | - |
| Fan speed | 30% | - | | | |
| Travel speed | 200 | mm/s | | | |
| Initial Travel speed | 130 | mm/s | | | |
| Build plate adhesion type | Brim | - | | | |

## 4. Discussion

The main contribution of this study is the systematic comparison of different manufacturing routes using the same complex pneumatic architecture and the same evaluation criteria. Previous studies have demonstrated the feasibility of direct and monolithic fabrication of soft pneumatic actuators using additive manufacturing [20], [21] [26]. In the present work, the emphasis is placed on process selection and development. Geometric fidelity, flexibility, structural integrity, airtightness, manufacturing complexity, and post-processing requirements were considered together, while the dominant failure mechanism of each process was used to determine whether further optimization was justified. This distinction between inherent process limitations and correctable manufacturing defects provides a practical basis for selecting manufacturing routes for complex soft pneumatic actuators.

The comparison revealed different trade-offs for each process. Heat-shrink forming offered a rapid and relatively simple fabrication route, but the available shrink ratio limited conformity to deep recessed features. Improvements in mold material, heating, and demolding reduced secondary problems but did not overcome this geometric limitation. Casting provided excellent material compliance, but the combination of narrow cavities, thin walls, and multiple bonded components increased the likelihood of incomplete filling, internal defects, and leakage. For the investigated geometry, increasing the wall thickness improved fabrication quality but also increased stiffness and reduced flexibility. Powder-based additive manufacturing provided greater geometric freedom, yet residual powder could not be removed reliably from narrow enclosed passages. DLP reproduced the complex geometry successfully, but the investigated resin system provided insufficient deformation for the intended actuator behavior and required several post-processing steps, including drainage, washing, vent-hole sealing, and post-curing. These results show that geometric capability alone is not sufficient to determine the suitability of a manufacturing process for a soft pneumatic actuator.

FDM was retained because its main limitations could be reduced through process development without simplifying the actuator architecture. The initial process was affected by unstable TPU feeding, moisture-related defects, incomplete deposition, weak bonding, support-related damage, and leakage. Unlike the dominant constraints encountered with the other routes, these problems were mainly process-dependent. They could therefore be addressed through material conditioning, extrusion stabilization, support-free design, controlled deposition, wall configuration, and cooling. FDM was not selected because it was inherently free of manufacturing difficulties, but because its principal defects could be controlled while preserving the required geometry, flexibility, and enclosed pneumatic structure. For the architecture examined here, this balance provided the clearest answer to the research question.

The FDM results also show that airtightness depends on the manufacturing process as a whole rather than on a single printing parameter. Filament condition influences material flow before deposition begins, while feeding stability, temperature, speed, flow rate, cooling, and wall configuration affect the continuity and bonding of the deposited material. Moisture control is a clear example. TPU absorbs moisture from the surrounding environment, and the observed sensitivity of extrusion stability and wall quality to filament condition led to drying and humidity control before the main printing parameters were optimized. During extrusion, absorbed moisture can vaporize and disturb melt flow, promoting pores and weak interfaces between deposited paths. Material conditioning should therefore be established before temperature, speed, or flow rate are optimized for airtight structures.

The wall-line experiments provided one of the most important findings of the study. A 0.96-mm wall formed from three 0.32-mm extrusion lines achieved better pneumatic sealing than a thicker 1.6-mm wall formed from two 0.8-mm lines. The experimental evidence therefore shows that nominal wall thickness alone does not determine pneumatic integrity. The number, width, and arrangement of the extrusion paths must also be considered. One possible explanation is that several narrower paths reduce the likelihood that local deposition defects align to form a continuous leakage path through the wall. Because these leakage paths were not directly visualized, this explanation remains a proposed mechanism rather than a confirmed failure mode. The measured result nevertheless identifies extrusion-path architecture as an important manufacturing variable and suggests that wall count and line width should be reported alongside nominal wall thickness in FDM-manufactured pneumatic actuators.

Support-free manufacturing produced a related design insight. Soluble and TPU supports could be introduced during printing, but removing them from enclosed regions either damaged thin walls or left defects in locations with limited access. Under the investigated material, printer, and geometric conditions, overhangs below approximately 30° were more likely to sag or form weak regions. The 30° value should therefore be regarded as specific to the present process rather than as a universal FDM limit. The broader finding is that internal accessibility should be considered during actuator design. Where post-processing access is limited, self-supporting geometry is preferable to relying on the later removal of internal supports.

The comparison also shows that manufacturing complexity cannot be judged simply by the number of fabrication steps. Heat-shrink forming was relatively simple but could not reproduce the required geometry. Casting used accessible materials but required multi-part tooling and bonded assembly. Powder-based methods avoided conventional support structures during printing but transferred the difficulty to powder removal after fabrication. DLP provided high geometric fidelity but required extensive material handling and post-processing. FDM required substantial process control during development, but once stabilized it could produce the integrated pneumatic geometry with limited secondary assembly and without inaccessible internal support material. The most suitable process was therefore the one whose limitations could be controlled without compromising the functional requirements of the actuator.

The study supports a failure-driven approach to manufacturing complex soft pneumatic actuators. Process selection should consider not only whether the required geometry can be produced, but also whether the dominant manufacturing limitations can be controlled while preserving flexibility, structural integrity, airtightness, and internal accessibility. Under the conditions investigated here, FDM provided the best balance of these requirements because its principal defects could be progressively reduced through process development. More broadly, the results show that stable material delivery, support-free design, controlled deposition, and extrusion-path architecture are central considerations when developing complex airtight pneumatic structures.

## 5. Conclusion

This study provides a systematic basis for manufacturing complex soft pneumatic actuators by linking process selection to geometric fidelity, flexibility, structural integrity, airtightness, internal accessibility, and post-processing requirements. A central outcome is the distinction between inherent process limitations and correctable manufacturing defects, allowing unsuitable routes to be identified early and viable processes to be developed more effectively. For FDM, successful pneumatic fabrication required coordinated control of material condition, extrusion stability, deposition quality, wall architecture, and support-free geometry. Notably, the results showed that airtightness depends on extrusion-path configuration rather than wall thickness alone. These findings support a design-for-manufacturing approach in which actuator geometry and fabrication constraints are considered together from the outset.

## 6. Acknowledgments

The author would like to express sincere gratitude to Dr. Marc Doumit for his valuable guidance, constructive advice, and continuous support throughout this research.

## 7. Funding

This research was funded by Dr. Marc Doumit and the University of Ottawa.